# Fast calculation of correlations in recognition systems

P. Dourbal[1] and M. Pekker[2]

[1] Dourbal Electric, Inc. 10 Schalk's Crossing Rd., St. 501-295, Princeton Jct., NJ 08536
paul@dourbalelectric.com

[2] The George Washington University Science & Engineering Hall 3550, 800 22nd Street, Northwest Washington, DC 20052, pekkerm@gmail.com

**Abstract**

Computationally efficient classification system architecture is proposed. It utilizes fast tensor-vector multiplication algorithm [1] to apply linear operators upon input signals. The approach is applicable to wide variety of recognition system architectures ranging from single stage matched filter bank classifiers to complex neural networks with multiple layers [2].

**Introduction**

Any signal recognition method requires forming a current observation vector $\vec{x}$ from a continuous information flow, comparing its distances to vectors from a set of templates $\vec{r}_\kappa$, ($k$

=1,2,.. K), and finally attributing the vector $\vec{x}$ to the closest template vector (Fig. 1) [3-6].

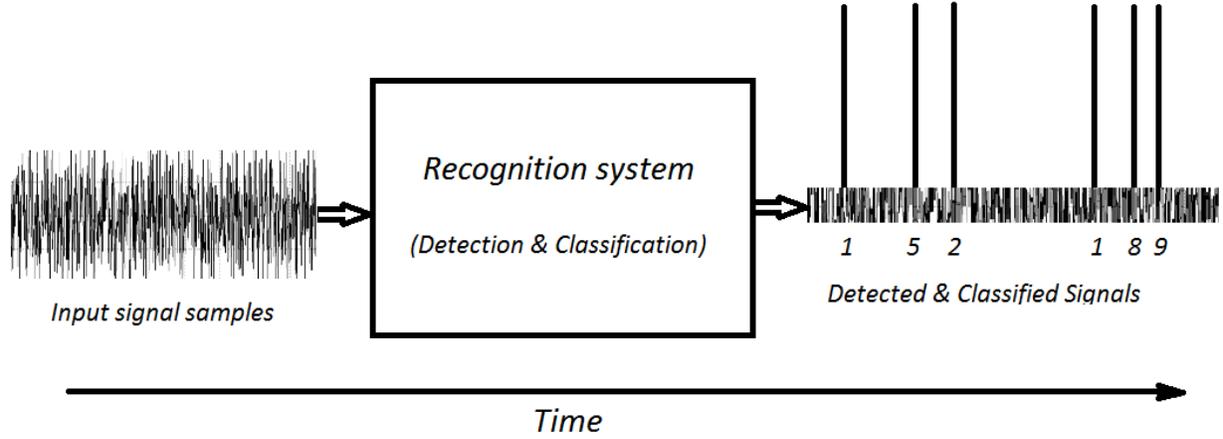

Fig 1. Signal detection and classification in recognition systems. Left – continuous-in-time input signal, right - detected and recognized signal. Numbers on the figure correspond to the numbers of templates present in the input signal and detected and classified by a recognition system.

The distance from the observed signal to the template $\kappa$ is:

$$d_k = \sum_{i=1}^{m}(x_i + r_{k,i})^2 = \sum_{i=1}^{m}x_i^2 + \sum_{i=1}^{m}r_{k,i}^2 - 2\sum_{i=1}^{m}x_i r_{k,i} = 2(1 - c_k) \qquad (1)$$

Here m is an observation and template vector length. We assume that all observation and template vectors are normalized so that: $\sum_{i=1}^{m}x_i^2 = 1$, $\sum_{i=1}^{m}r_{k,i}^2 = 1$.

A value $c_k$, given by

$$c_k = \sum_{i=1}^{m} x_i r_{ik} \qquad (2)$$

is called the cross-correlation coefficient [4]. The greater the value of $c_k$, the closer the vector $\vec{x}$ is to the template $\vec{r}_k$. A set of correlations $c_k$, ($k$ =1,2,.. K), is called the correlation vector. According to (2), it can be represented by the product of the template matrix and the observation vector:

$$\vec{c} = [R] \cdot \vec{x}' \qquad (3)$$

The dimensions of the matrix $[R]$ are $m \times K$ since its rows are template vectors of length K. Training the recognition system involves forming the set of template vectors $\vec{r}_k$ corresponding to the classes of objects to be recognized. The training process is independent of the recognition algorithm.

The effectiveness of an object recognition system is determined by the resources required to compute the correlation vector $\vec{c}$ for each new observation vector $\vec{x}_n$ obtained by shifting the preceding vector $\vec{x}_{n-1}$ by one position, such that

$$\vec{x}_{n-1} = \begin{Bmatrix} a_1 \\ a_2 \\ a \\ \ldots \\ \ldots \\ a_{m-2} \\ a_{m-1} \\ a_m \end{Bmatrix}, \quad \vec{x}_n = \begin{Bmatrix} a_2 \\ a_3 \\ a_4 \\ \ldots \\ \ldots \\ a_{m-1} \\ a_m \\ a_{m+1} \end{Bmatrix}, \quad \vec{x}_{n+1} = \begin{Bmatrix} a_3 \\ a_4 \\ a_5 \\ \ldots \\ \ldots \\ a_m \\ a_{m+1} \\ a_{m+2} \end{Bmatrix} \qquad (4)$$

**Direct method of calculation of correlations**

The direct method of calculation of correlations [7] according to formula (3) requires *m-1* addition and *m* multiplication operations to calculate one correlation coefficient $c_k$. In the direct method for every new vector $\vec{x}_n$ it is necessary to perform a complete calculation of the matrix by a vector product, so the number of additions and multiplications required to obtain each subsequent correlation vector $\vec{c}_n$ is:

$$M_+ \approx M_* = K \cdot m \qquad (5)$$

In some cases template matrices can be converted to the Tensor Train [8] format reducing the number of elementary operations in the matrix-vector multiplications by about 7 times [2].

## Correlation calculation using the Dourbal algorithm

The Dourbal algorithm [1] is a method for the automatic synthesis of fast linear transformation algorithms for rapid matrix by vector or tensor by vector products. This algorithm allows for significant reductions in computational complexity and therefore decreases the computational resources required to obtain a correlation vector $\vec{c}$ in the direct method of correlation computation as well as in the Viterbi and other methods based on dynamic programming. The efficiency of the algorithms for correlation calculation automatically synthesized with the Dourbal method is defined by the number of significant figures after the decimal point in components of the template vectors forming the correlation matrix $[R]_{K,m}$ elements. Figure 2 shows the relationship between the number of elementary operations of multiplication and addition required to multiply the matrix $[R]_{K,m}$ by the vector $\vec{x}$ and the size of the matrix $[R]_{K,m}$ for the standard method and for the Dourbal algorithm. The curves on figure 2 correspond to the number of significant figures in the representation of the matrix $[R]_{K,m}$. Here it must be mentioned that in recognition systems the number of significant figures is usually limited and usually doesn't exceed three [9].

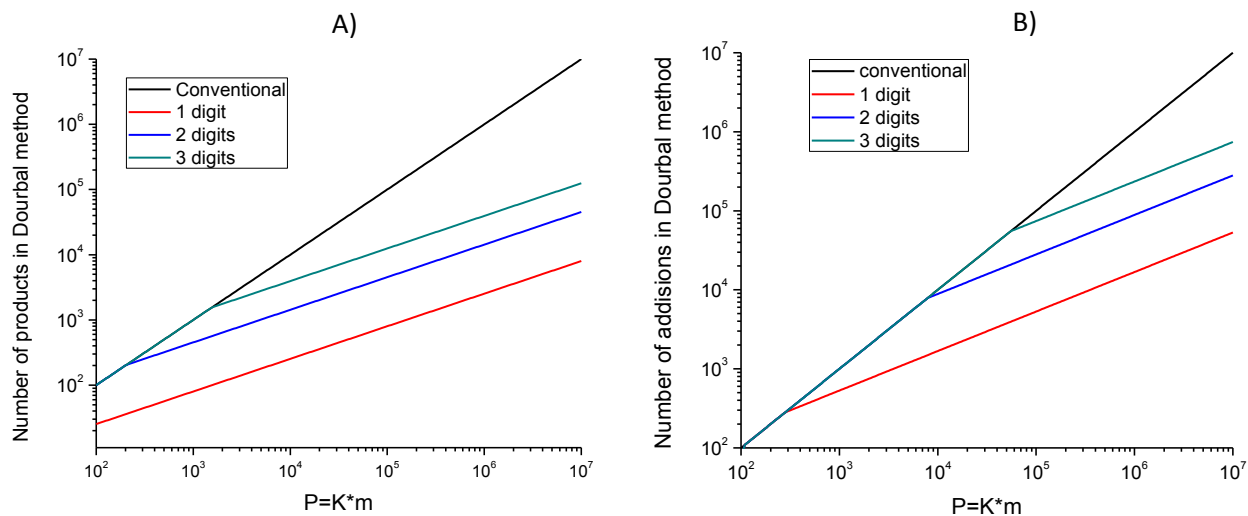

Figure 2. Dependence of the number of elementary operations to multiply matrix $[R]_{K,m}$ by vector $\vec{x}$ on $P = K*m$. (A) - products, (B) – additions.

## Calculation of correlations for a continuous signal

Usually in signal recognition tasks the template or correlation matrix $[R]_{K,m}$ is not changing. Meanwhile, every new observation vector $\vec{x}$ differs from a previous observation vector by one position (4) [10]. In those cases, for some types of template waveforms, the Viterbi algorithm or its modifications are utilized. They make it possible to decrease the dimension m of the matrix $[R]_{K,m}$. In the Viterbi algorithm, with the arrival of each new sample of the input signal, intermediate vectors are calculated. Later those vectors are used to obtain correlation vectors $\vec{c}$ [9-10]. Accordingly, the Viterbi algorithm makes it possible to reduce the number of addition and multiplication operations required to calculate a correlation vector $\vec{c}$ to:

$$M_+ \approx M_* = \alpha_v \cdot P, \qquad (6)$$

where $\alpha_v < 1$. In [11] it was shown that for a continuous input signal, the number of multiplication operations can't be decreased by a factor of more than 2, and the number of additions remains practically unchanged:

$$\alpha_v^{(*)} = \begin{cases} 1 & \text{for } P < 48 \\ 1/2 & \text{for } P > 48 \end{cases} \qquad (7)$$

$$\alpha_v^{(+)} = \begin{cases} 1 - 1/\sqrt{P} & \text{for } P < 48 \\ 1 - \sqrt{48} & \text{for } P > 48 \end{cases} \qquad (8)$$

It is known that the Viterbi algorithm is applicable only for Markovian processes of the first order [9,10]. Meanwhile, the Dourbal algorithm is independent of the form of the statistical distribution function of a stochastic process.

## Calculation of correlations for a continuous signal using the Dourbal algorithm

The Dourbal algorithm stores and uses the intermediate results of all elementary summation and multiplication operations. Therefore, for a continuous input signal, where each new observation vector is obtained by shifting the previous vector by one position (4), the numbers of elementary summation and multiplication operations required to calculate a subsequent correlation vector drop significantly. Fig. 3 shows the numbers of elementary multiplication and summation operations required for calculation of one correlation vector for a continuous input signal as a function of the matrix $[R]_{K,m}$ size $P = K \cdot m$.

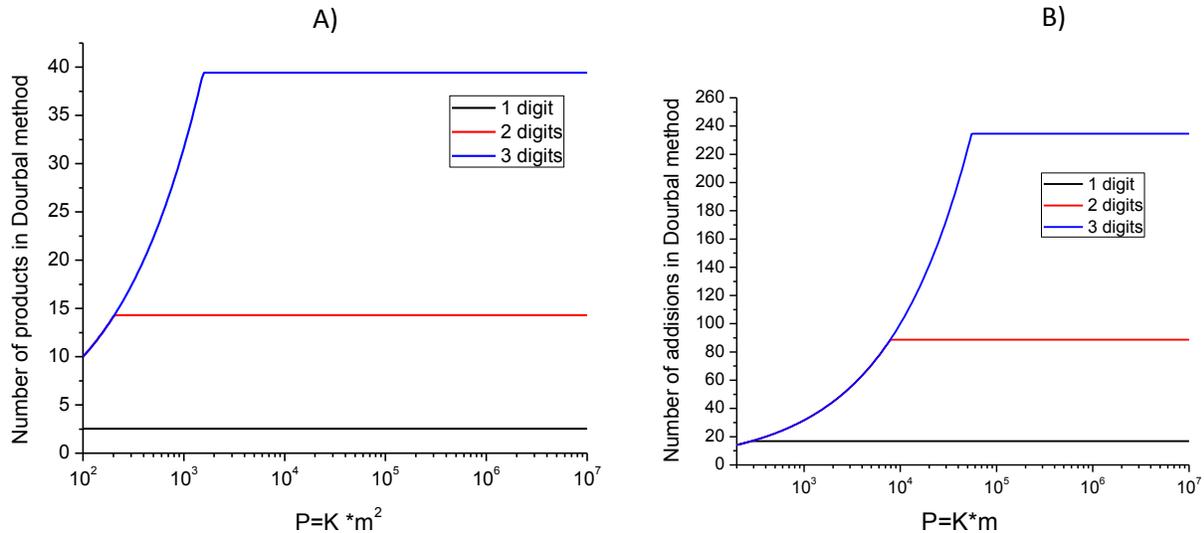

Figure 3. Dependence of the number of elementary operations for correlation vector calculation obtained by multiplication of matrix $[R]_{K,m}$ by a vector $\vec{x}$ as a function of the matrix size $P = K \cdot m$ for a continuous input signal. (A) - products, (B) – additions.